\PassOptionsToPackage{numbers,sort&compress}{natbib}
\documentclass[sigconf,nonacm]{acmart}

\usepackage[utf8]{inputenc}
\usepackage[T1]{fontenc}
\usepackage{microtype}
\usepackage{booktabs}
\usepackage{tabularx}
\usepackage{amsmath}
\usepackage{xcolor}
\usepackage{url}

\setcopyright{none}
\hypersetup{
  hidelinks,
  pdftitle={Generation Provenance Before Behavior Attribution: Auditing Synthetic Speech Research Objects},
  pdfauthor={Sidi Chang; Peiying Zhu}
}
\providecommand{\doi}[1]{\url{https://doi.org/#1}}

\newcommand{\obj}{\mathcal{O}}

\begin{document}
\acmshorttitle{Generation Provenance Before Behavior Attribution}
\acmshortauthors{Chang \& Zhu}
\pagestyle{acmstyle}
\thispagestyle{firstpage}

\twocolumn[
\begin{@twocolumnfalse}
\begin{center}
{\LARGE\bfseries Generation Provenance Before Behavior Attribution: Auditing Synthetic Speech Research Objects\par}
\vspace{1.0em}

\begin{tabular}{@{}c@{\hspace{2em}}c@{}}
{\large Sidi Chang\textsuperscript{*\,$\dagger$}} &
{\large Peiying Zhu\textsuperscript{*}} \tabularnewline
{\small schang@blossomai.co} & {\small peiying@blossomai.co} \tabularnewline
{\small Blossom AI Labs} & {\small Blossom AI} \tabularnewline
{\small Tokyo, Japan} & {\small San Francisco, CA, USA}
\end{tabular}
\vspace{1.2em}
\end{center}

\noindent\parbox{\textwidth}{
\small
\noindent\textbf{Abstract}\\[3pt]
Attributing model behavior to synthetic training data requires knowing what produced each training item before estimating what that item caused. A waveform--label pair does not preserve this knowledge. We propose a generation-provenance substrate in which a synthetic research object binds source specification, generated content, waveform, target, fact requirements, quality signals, review lineage, and immutable manifest identity. Producer and selection mechanism determine evidentiary meaning; storage location and variable name do not. We audit this substrate in a private Japanese care-handoff pipeline. A 113-asset review population contains 1.552 hours of synthetic speech across six scenario families; all items have linked audio, transcripts, candidate notes, and fact checklists, but human evidence is selective and source-specific. Two faithful-only manifests are scenario-seed-disjoint and immutably versioned, while exact upstream attribution remains blocked by floating generator aliases, missing per-clip TTS and code stamps, and an unversioned checking prompt. We argue that generation provenance is necessary but not sufficient for behavior attribution: it defines the candidate causal graph and audit units, whereas contributive attribution still requires frozen training runs and intervention or influence evidence. The paper contributes a compact provenance contract, an audit protocol, and a bounded case study for synthetic-data attribution; controlled research access may be offered, but we do not claim causal training-data attribution, clinical validity, or unrestricted public release.
}

\vspace{1.5em}
\end{@twocolumnfalse}
]

\renewcommand{\thefootnote}{\fnsymbol{footnote}}
\footnotetext[1]{Both authors contributed equally to this research.}
\footnotetext[2]{Corresponding author.}
\renewcommand{\thefootnote}{\arabic{footnote}}
\setcounter{footnote}{0}

\section{Why synthetic attribution starts upstream}

Data attribution asks which training data shaped a model output or action. Synthetic data complicates this question because a nominal ``example'' is itself the output of upstream models, prompts, source specifications, filters, speech renderers, and selection policies. Two identical waveform--target pairs can have different upstream sources, rights, error mechanisms, and reproducibility. Conversely, two regenerated assets may share a semantic seed while differing byte-for-byte. Treating the final pair as the only attribution unit discards the structure needed for legal, scientific, and operational audits.

Existing documentation frameworks emphasize composition, collection, intended use, and limitations \cite{bender-friedman-2018-data,gebru-etal-2021-datasheets,pushkarna-etal-2022-data-cards}; large-scale audits show that licenses and derivation chains are frequently missing \cite{longpre-etal-2024-provenance}. Synthetic-data research further shows why quality, diversity, and task structure must be measured rather than inferred from the generator \cite{chim-etal-2025-evaluating,wang-etal-2025-unifilter,yun-etal-2025-price}. We connect these concerns to behavior attribution.

Our central claim is a prerequisite, not an attribution algorithm. Let $z$ denote an upstream source specification, $x$ a generated training item, $m$ a manifest, $r$ a training run, and $b$ an observed behavior. A behavior-attribution claim about $z$ is auditable only if the system preserves versioned relations
\begin{equation}
z \longrightarrow x \longrightarrow m \longrightarrow r \longrightarrow b,
\end{equation}
with explicit producer, transformation, and identity semantics. Preserving this path does not prove that $z$ caused $b$; it identifies the candidate path on which contributive or corroborative attribution can operate. If an edge is missing, multiple upstream histories are observationally equivalent at the stored record.

For a minimal counterexample, consider two histories that store the same $(x,m,r,b)$. In history A, source $z$ generates $x$; in history B, a different source $z'$ generates the same $x$ and $z$ is attached later as an untyped label. A query attributing $b$ to $z$ should include the training item in A but exclude it in B, yet any method observing only $(x,m,r,b)$ receives identical input. Both histories are byte-identical under any schema that lacks a typed \texttt{generated-from} edge. No downstream attribution score can recover the missing typed edge.

We instantiate this idea in a synthetic Japanese care-handoff pipeline. The contribution is threefold: an aligned research object for synthetic speech, a five-pass provenance audit, and an empirical account of which later attribution questions current records enable or block. The case is deliberately bounded to provenance and does not evaluate compact-model capability or judge calibration.

\section{Provenance substrate}

For item $i$, we define
\begin{equation}
\obj_i=\langle I_i,S_i,R_i,A_i,Y_i,F_i,M_i,Q_i,H_i,P_i\rangle,
\end{equation}
where $I$ is stable identity; $S$ source scenario semantics; $R$ spoken realization; $A$ audio; $Y$ structured target; $F$ fact requirements; $M$ dialogue and acoustic metadata; $Q$ automatic alerts and planted defects; $H$ human-review lineage; and $P$ version provenance. Missing components remain explicit rather than being silently defaulted.

The audited generation graph contains: 32 hand-authored scenario seeds; generated six-field notes and fact checklists; code validation; generated spoken Japanese; a model-based extractiveness check; and speech rendering with multiple voices. Controlled omissions and fabrications are planted only after faithful generation. An advisory ASR-consistency alert is also computed post hoc. Human review is selective downstream evidence, not a generation stage.

This graph motivates five invariants. \textbf{Identity alignment} links the scenario, transcript, note, checklist, audio, and review records. \textbf{Semantic anchoring} retains the intended facts and events. \textbf{Typed signal semantics} distinguishes a generator key, a detector output, an advisory alert, and a human judgment. \textbf{Split safety} assigns semantic seeds rather than clips across partitions. \textbf{Version identity} pins every claim to an immutable manifest instead of a mutable project name.

These invariants define different attribution units. A scenario seed supports analysis at the semantic-source level; a clip supports acoustic analysis; a fact supports output-level corroboration; a planted defect supports intervention audits; and a manifest row links an item to a training run. Collapsing these units into one ``data point'' creates attractive but ambiguous attribution scores.

\section{Five-pass audit}

Each audit finding is stored as $\langle$population, grain, check, finding, risk, supported claim$\rangle$. The passes are: (1) resolve keys and required fields; (2) join source, text, target, fact, and audio records; (3) type every quality signal by producer and operational effect; (4) test manifest counts, seed leakage, and version relations; and (5) inventory generation metadata, access, and license scope.

We keep three populations separate. Factory populations contain faithful and deliberately planted records. The review population contains 113 assets selected into internal quality workflows. Faithful-only training manifests exclude planted rows. Counts from one population are never used as a denominator for another.

\begin{table*}[t]
\centering
\small
\begin{tabularx}{\textwidth}{@{}p{0.14\textwidth}X X X@{}}
\toprule
Record & Observed evidence & Supports & Does not support \\
\midrule
Assets & 113 IDs and linked audio; 1.552 h; 32 seeds; one empty Focus & cross-artifact identity and observed coverage & target correctness or representativeness \\
Realism & 75 final-rubric items; one label each; 58 usable, 17 minor; 72 natural & selected input/audio plausibility & paired-target validity or agreement \\
Correction & 12 notes; one rater each; eight changed & exploratory target defects & certified gold or error prevalence \\
Error cards & selective final records with source-linked dispositions and human additions; sparse overlap & review of displayed plant keys & detector precision/recall \\
\bottomrule
\end{tabularx}
\caption{Human evidence is source- and workflow-specific. ``Plant'' and ``detector output'' are not interchangeable provenance labels.}
\label{tab:evidence}
\end{table*}

The asset pass finds 113 unique target IDs, 113 linked media assets, and nonempty transcripts, candidate notes, and fact checklists for all items; one note has an empty Focus field. The assets cover six scenario families and two care settings and total 5,586.88 seconds of 24-kHz mono audio. These are properties of the selected review population, not the factory or released training data.

Table~\ref{tab:evidence} shows why quality cannot be a single column. An ASR alert is an uncalibrated consistency tripwire and is retained in faithful training data. A planted key records an intended intervention and is excluded from faithful training. Realism labels concern spoken input and audio. Corrections concern candidate targets. Error-card judgments dispose of displayed plant proposals; they are not independent judge detections. Producer and selection mechanism determine meaning, not a table or variable name.

\section{Identifiable claims}

The first faithful manifest contains 221 clips split 139/43/39 across train/development/held-out and 32 seeds split 20/6/6. A later manifest contains 313 clips split 184/68/61 and 58 seeds split 35/12/11. Both have zero cross-partition seed overlap; the first is an exact subset of the second, which adds 92 faithful rows. Each is pinned to an immutable revision, denoted $h_A$ and $h_B$ during double-blind review. These records establish item identity and partition lineage. They do not turn version movement into a causal scaling experiment because both training and held-out composition change.

Existing experiment records partially close the downstream path. Two adaptation runs consume the same 182 train-plus-development rows from $h_A$, and each stored output set contains exactly the 39 held-out IDs. Thus $m\rightarrow r\rightarrow b$ is auditable at run and item identity, even though no per-training-example influence score exists. This distinction matters: exact manifest-to-output lineage supports attribution studies without itself establishing which training item caused a behavior.

The graph already enables useful attribution research. One can compare inclusion of ASR-alerted faithful rows; localize errors to target generation, spoken realization, rendering, or evaluation; stratify behavior by scenario family, seed, dialogue mode, and required fact; or test whether a behavior traces to planted interventions while keeping plant keys separate from detector outputs. These analyses require training-run and behavior records to be joined to the manifest, but the candidate units and relations now exist.

Exact upstream attribution is not yet possible. Generator names are floating aliases rather than dated snapshots; clips lack per-item TTS-backend and generating-code stamps; one model-based checking prompt has no version constant; a union artifact masks some seeds; and the later shard lacks an ASR-gate record. Sparse independent review also blocks population-level validity claims. These gaps do not erase the stored assets, but they create equivalence classes of possible generation histories.

\section{Implications and limitations}

For synthetic data, behavior-attribution systems should preserve a minimal contract: content and item hashes; source seed and selection policy; generator, prompt, temperature, code, and renderer versions; pre- and post-filter status; immutable manifest membership; training-run identity; and behavior-output identity. Typed edges should distinguish \texttt{generated-from}, \texttt{selected-into}, \texttt{trained-on}, \texttt{detected-by}, \texttt{reviewed-as}, and \texttt{matched-to}. Many-to-many relations should be retained rather than forced into one source label.

This contract is commercially relevant without making the data a product claim: it allows a future researcher to purchase or access a fixed object whose derivation, eligible uses, and experimental units are inspectable. The corpus is wholly synthetic and contains no patient recordings or health-record data. The paper uses de-identified aggregates from an existing lawful internal quality-review workflow and does not analyze reviewer identities or participant outcomes. Audio, transcripts, targets, fact checklists, metadata, and provenance may be offered to qualified researchers under controlled access, a data-use agreement, security requirements, and applicable third-party terms. Academic and commercial access may use different licenses and fee schedules. No unrestricted public availability or reuse is claimed.

An aggregate reproducibility package is available from the authors on request. It contains a frozen aggregate count sheet, a standard-library verification script, an executed audit notebook, a paper-to-number claim crosswalk, and a machine-readable provenance-object schema with a wholly synthetic fixture. The package reproduces displayed arithmetic, count identities, and manifest relations; it cannot replay the private extraction or substitute for access to the controlled corpus.

The study audits one synthetic clinical-speech pipeline with 1.552 reviewed hours. Review is selective, has one independent realism label per item, and provides only sparse overlapping error-card review. Generated notes and fact lists are not clinical gold. The new manifest was not created as a controlled scaling intervention, and no public corpus license or fee schedule has yet been selected. Most importantly, the paper does not compute the causal effect of any training item on model behavior. Generation provenance narrows what attribution can mean and makes later analysis auditable; it does not replace influence estimation, counterfactual retraining, or corroborative evidence.

\section{Conclusion}

Synthetic training data has two histories: how an item was generated and how it affected a model. Behavior attribution becomes scientifically ambiguous when the first history is discarded. Our aligned object and audit preserve stable units, typed evidence, seed-safe partitions, and immutable manifests while exposing missing upstream stamps and access terms. The result is a prerequisite for attribution, not attribution itself: a versioned path from source specification to training and behavior on which causal or corroborative methods can make testable claims.

\bibliographystyle{plain}
\bibliography{references}
\end{document}